\documentclass[dvipsnames]{article} %
\usepackage{template_conference}

\usepackage{booktabs}
\usepackage{graphicx}
\usepackage{enumitem}
\usepackage{wrapfig}
\usepackage{algorithm}
\usepackage{algpseudocode}
\usepackage[utf8]{inputenc}
\DeclareUnicodeCharacter{FF5E}{\textasciitilde}

\usepackage{adjustbox}
\usepackage{amsmath}
\usepackage{colortbl}
\usepackage[utf8]{inputenc}
\definecolor{lightgray}{rgb}{0.9,0.9,0.9}
\usepackage{caption}
\usepackage{subcaption}
\usepackage[svgnames]{xcolor}
\usepackage{setspace}
\usepackage{url}
\usepackage{multirow}
\usepackage{colortbl}
\usepackage{tabularx}
\usepackage{blindtext}
\usepackage{pgfplots}
\pgfplotsset{compat=1.18} 
\usepackage{tikz}
\usetikzlibrary{er,positioning}
\usepackage{makecell}
\usepackage{tipa}
\usepackage{siunitx}
\usepackage{tocloft}
\usepackage{listings}
\usepackage{adjustbox}
\usepackage{xurl}
\usepackage{rotating}
\usepackage[normalem]{ulem}
\usepackage{multicol}
\usepackage{titlesec}
\usepackage{tcolorbox}
\tcbuselibrary{breakable,theorems}
\usepackage{transparent}
\usepackage{mathtools}

\usepackage{amsthm}

\usepackage{amsfonts}
\usepackage{etoc}
\usepackage{color}
\usepackage{autobreak}
\usepackage{soul}
\usepackage{amssymb}
\usepackage{bbding}
\usepackage[misc]{ifsym}

\usepackage{amsmath}
\usepackage{pifont}
\newcommand{\cmark}{\ding{51}}
\newcommand{\xmark}{\ding{55}}

\definecolor{lightblue}{RGB}{232,242,255}
\definecolor{stdblue}{RGB}{0,114,178}

\newtheorem{proposition}{Proposition}

\newtcolorbox{codeblock}{
  colback=gray!10,   
  colframe=gray!50,  
  boxrule=0.5mm,     
  arc=2mm,           
  left=5pt,          
  top=5pt,          
  bottom=5pt,        
}

\newtcbtheorem{definition}{Definition}{
  enhanced,
  colback=cyan!5!white,
  colframe=cyan!45!blue!60,       
  fonttitle=\bfseries,            
  fontupper=\normalfont,          
  arc=4mm,                        
  boxrule=1pt,
  drop shadow southwest,
  attach boxed title to top left={xshift=6mm, yshift=-2mm},
  boxed title style={
    colback=cyan,
    size=small,
    arc=2mm,                  
    boxrule=0.8pt,
    left=2mm, right=2mm
  },
  before skip=10pt,
  after skip=10pt,
  left=5mm,
  right=5mm,
  top=3mm,
  bottom=3mm
}{cor}

\useunder{\uline}{\ul}{}

\usepackage{amsmath,amsfonts,bm}

\def\eqref#1{equation~\ref{#1}}

\def\1{\bm{1}}

\DeclareMathAlphabet{\mathsfit}{\encodingdefault}{\sfdefault}{m}{sl}
\SetMathAlphabet{\mathsfit}{bold}{\encodingdefault}{\sfdefault}{bx}{n}

\renewcommand{\ghlink}{YourGithubLink}

\newcommand*\justify{%
  \fontdimen2\font=0.4em
  \fontdimen3\font=0.2em
  \fontdimen4\font=0.1em
  \fontdimen7\font=0.1em
  \hyphenchar\font=`\-
}

\renewcommand{\texttt}[1]{%
  \begingroup
  \ttfamily
  \begingroup\lccode`~=`/\lowercase{\endgroup\def~}{/\discretionary{}{}{}}%
  \begingroup\lccode`~=`[\lowercase{\endgroup\def~}{[\discretionary{}{}{}}%
  \begingroup\lccode`~=`.\lowercase{\endgroup\def~}{.\discretionary{}{}{}}%
  \catcode`/=\active\catcode`[=\active\catcode`.=\active
  \justify\scantokens{#1\noexpand}%
  \endgroup
}

\definecolor{blueviolet}{RGB}{138,43,226}
\newtcolorbox{insightblock}{
  colback=blueviolet!5,
  colframe=blueviolet!50!black!50!,
  boxrule=0.5mm,       
  arc=2mm,             
  left=0pt,            
  right=8pt,           
  top=8pt,             
  bottom=8pt,          
}

\newtcolorbox{abstractbox}{
    colback=blue!5!white,     
    frame empty,              
    boxrule=1pt,              
    arc=4mm,                  
    left=8pt,                 
    right=8pt,                
    top=8pt,                  
    bottom=8pt,                
    opacityback=0.9
}

\newtcolorbox{examplebox}[1][]{
    breakable,
    colback=black!4,
    colframe=black!35,
    boxrule=0pt,
    sharp corners,
    boxsep=0pt,
    left=8pt,
    right=7pt,
    top=6pt,
    bottom=6pt,
    before upper={
        \if\relax\detokenize{#1}\relax
        \else
            {\small\bfseries\color{black!85} #1\par}
            \vspace{3pt}
        \fi
    },
    before skip=0.8\baselineskip,
    after skip=0.8\baselineskip,
    pad at break*=1.2mm
}

\title{Scaling Action-Free Video Latent Prediction for Transferable Representations in End-to-End Autonomous Driving}

\author{
\footnotesize
\textbf{Xingtai~Gui}\textsuperscript{1*\textdagger} \quad
\textbf{Xubin~Zhou}\textsuperscript{2*}
 \quad
\textbf{Yucheng~Zhou}\textsuperscript{1*} \quad
\textbf{Dongqian~Guo}\textsuperscript{1} \quad
\textbf{Jiahao~Gong}\textsuperscript{3} \quad 
\\[2pt]
\textbf{Feiyang~Tan}\textsuperscript{3} \quad
\textbf{Jianbing~Shen}\textsuperscript{1\textdaggerdbl} \quad
\\[2pt]
\scriptsize \textsuperscript{1}SKL-IOTSC, Department of AI, University of Macau\quad
\textsuperscript{2}Harbin Institute of Technology\quad
\textsuperscript{3}Afari Intelligent Drive
}

\begin{document}

\maketitle
\begingroup
\renewcommand{\thefootnote}{\fnsymbol{footnote}}
\footnotetext[1]{Equal contribution.}
\footnotetext[2]{Work done during the internship at Afari Intelligent Drive.}
\footnotetext[3]{Corresponding author.}
\endgroup

\begin{abstractbox}
\begin{center}
\textbf{\Large Abstract}
\end{center}
End-to-end autonomous driving policies require effective visual representations that capture both spatial scene structure and temporal dynamics. Video latent prediction has emerged as an effective paradigm for learning spatiotemporal visual representations, yet it remains unclear whether scaling this objective with action-free driving videos can yield representations that benefit end-to-end planning. We study action-free video latent prediction as a scalable pretraining objective and introduce ScaleDrive, a driving-oriented framework pretrained exclusively on large-scale internet open-world driving videos, without action annotations or benchmark-specific data. We keep the pretrained video encoder frozen during downstream end-to-end planning training to isolate the transferability of representations learned from internet driving videos. Ego-motion and depth probes show that scaling action-free video pretraining progressively strengthens the temporal and geometric information encoded in the frozen representation. We systematically study the visual representation transfer across different end-to-end driving policies and closed-loop benchmarks, where ScaleDrive consistently improves downstream performance and demonstrates that scaling action-free video latent prediction yields representations that generalize across both planning policy architectures and evaluation scenarios and protocols. 

Project: \url{https://um-lab.github.io/ScaleDrive}
\end{abstractbox}



\section{Introduction}
Learning generalizable visual representations is a central challenge for end-to-end autonomous driving. A strong representation should capture not only scene appearance, but also spatial structure, interactions among traffic participants, and temporal evolution. Existing driving representations are often learned through task-specific objectives, including object detection~\citep{bevdet, bevformer, petr}, online mapping~\citep{maptr,liu2024mgmap, maptrv2}, occupancy
prediction~\citep{occ3d, triformer, occformer}, and
vision-language-action modeling~\citep{wang2025alpamayo,
zhou2026opendrivevla, zhou2026autovla}. Although effective within their respective task formulations, these objectives primarily encourage representations to capture information required by predefined semantic, geometric, or planning targets, rather than broadly modeling the general structure and dynamics of the driving world.

\begin{figure}[t]
\centering
\includegraphics[width=\linewidth]{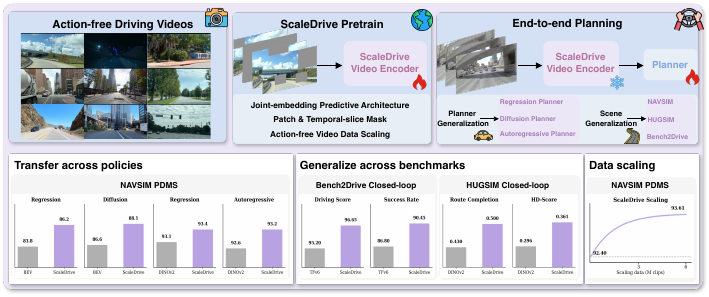}
\vspace{-7mm}
\caption{
ScaleDrive learns transferable driving representations through self-supervised predictive pretraining on large-scale raw driving videos. The pretrained encoder is kept frozen and transferred across heterogeneous end-to-end planners and closed-loop driving environments, consistently improving downstream planning performance.
}
\label{fig:intro}
\vspace{-4mm}
\end{figure}

Self-supervised video learning provides a complementary paradigm by learning directly from how visual scenes are structured and evolve over time, without requiring task-specific supervision~\citep{videomae, wang2024internvideo2, vjepa2}. In particular, video latent prediction learns spatiotemporal representations by predicting masked content in representation space~\citep{bardes2024revisiting, vjepa2}. This paradigm is particularly attractive for autonomous driving: raw driving videos naturally capture road geometry, traffic participants, interactions, and scene dynamics without requiring additional detection, mapping, occupancy, reasoning, or action annotations. They can therefore support large-scale representation pretraining without additional task-specific labeling. Although prior works have explored self-supervised pretraining from driving videos~\citep{drivex, vavim, drivejepa, gift}, an important question remains underexplored: \textit{does driving representation quality continue to improve as raw driving video pretraining data scales? More importantly, do these representation gains translate into better downstream planning and transfer across different planning policies and driving scenarios?}

To study this question, we introduce \textbf{ScaleDrive}, a scalable self-supervised video representation learning framework for autonomous driving, as illustrated in Fig.~\ref{fig:intro}. ScaleDrive learns solely from raw driving videos, without action or task-specific supervision. It follows a joint-embedding predictive architecture (JEPA)~\citep{bardes2024revisiting, vjepa2}, where visible video regions are encoded by a context encoder, and the latent representations of masked regions are predicted under the supervision of targets produced by the momentum target encoder. To strengthen temporal modeling, we augment conventional patch-level masking with complete temporal-slice masking, encouraging the model to capture both spatial scene structure and temporal dynamics. For downstream evaluation, we freeze the pretrained video encoder and train only the planning module, helping isolate the contribution of the learned representation from downstream representation adaptation.

Our experiments reveal a consistent representation scaling trend. As pretraining data increases, downstream planning performance improves while the visual encoder remains frozen. Ego-motion and monocular depth probing further show that temporal-motion and spatial-geometric information becomes increasingly accessible from the learned representation. Together, these results indicate that scaling video latent prediction pretraining produces increasingly useful driving representations.

We further evaluate whether these gains transfer beyond a single downstream planning configuration. The same frozen ScaleDrive representation consistently improves regression, diffusion, and autoregressive planners, and remains effective across distinct closed-loop benchmarks including HUGSIM~\citep{zhou2025hugsim} and Bench2Drive~\citep{jia2024bench2drive}. These results show that scaling self-supervised predictive pretraining on raw driving videos can progressively strengthen transferable driving representations and improve downstream planning across heterogeneous policy architectures and driving environments.

The main contributions are summarized as follows:
\begin{itemize}[leftmargin=*]
\item We introduce \textbf{ScaleDrive}, a scalable self-supervised pretraining framework that learns transferable driving representations solely from large-scale raw driving videos, without action annotations or task-specific supervision.
\item We develop a \textbf{driving-oriented predictive pretraining strategy} that combines patch-level masking with temporal-slice masking to capture both spatial scene structure and temporal dynamics.
\item We conduct a \textbf{systematic study of representation scaling}. As video pretraining data increases, downstream planning performance improves, while ego-motion and depth probing reveal progressively stronger temporal-motion and spatial-geometric information in the frozen representation.
\item ScaleDrive achieves \textbf{competitive end-to-end planning performance with a frozen visual encoder} and consistently improves regression, diffusion, and autoregressive planners and generalizes across distinct benchmarks including NAVSIM, HUGSIM and Bench2Drive.
\end{itemize}

\section{Related Works}
\label{sec:related}
Existing studies on visual representation learning for autonomous driving span task-specific intermediate representations such as BEV and occupancy~\citep{bevformer, bevdet, occ3d, occformer}, generic foundation-model features from large-scale pretrained vision and vision-language models~\citep{oquab2023dinov2, cogvideox, wang2024qwen2, kirby2026driving, zhang2025epona, gui2026bridging, zhou2026autovla, jiang2026senna}, and video-based joint-embedding predictive learning~\citep{vjepa2, drivejepa, yang2026auto}. In parallel, scaling studies show that driving policies benefit from more action-supervised data~\citep{naumann2025data, zheng2024data, tian2026simscale, li2026drivevla}, and representation learning benefits from larger-scale data and models~\citep{kong2025largead, li2026stellar}. However, whether \emph{action-free} driving representations can be systematically scaled and transferred to end-to-end planning remains underexplored. The full version of related works can be found in Appendix~\ref{app:related}.

\section{ScaleDrive}

ScaleDrive learns transferable driving representations from action-free videos through self-supervised latent prediction. It builds upon the joint-embedding predictive architecture, using raw driving videos without trajectory or perception annotations for pretraining. After pretraining, the predictor is discarded and the video encoder is frozen for downstream end-to-end driving policies, allowing downstream performance to more directly reflect the quality of the learned representation. A key design challenge is that conventional patch-level prediction may exploit same-frame spatial redundancy without sufficiently modeling temporal dynamics. We therefore introduce complete temporal-slice masking to complement patch-level masking, encouraging prediction from temporally distributed visual context. We next provide the theoretical motivation for this design and describe the resulting spatiotemporal latent prediction framework.

\subsection{Dynamics-Guided Temporal Prediction}

We first analyze latent prediction for driving videos from a dynamic-system perspective. The purpose of this analysis is not to model the full complexity of real driving dynamics, but to characterize a sufficient condition under which patch-level prediction admits a same-frame shortcut and to explain why complete temporal masking provides a stronger temporal learning signal.

\paragraph{Driving dynamic system.}
A driving video can be viewed as a sequence of partial observations generated by an underlying dynamic scene state. Let
$\mathbf{s}_t \in \mathbb{R}^{d_s}$
denote the latent physical state of the driving scene, including ego motion, surrounding-agent states, and local scene geometry. Its temporal evolution and visual observation can be abstracted as
\begin{equation}
    \mathbf{s}_{t+1}
    =
    F(\mathbf{s}_t)+\mathbf{w}_t,
    \qquad
    \mathbf{y}_t
    =
    G(\mathbf{s}_t)+\mathbf{v}_t,
\end{equation}
where $\mathbf{w}_t$ and $\mathbf{v}_t$ denote process and observation noise, respectively, and $\mathbf{y}_t$ denotes a temporally indexed latent observation corresponding to the predictive-learning target. Here, $\mathbf{s}_t$ is an abstract physical latent introduced only for analysis and we do not assume that a neural encoder explicitly recovers this state.

\paragraph{Local linearization.}
Over a sufficiently short video clip, we analyze the local behavior of the nonlinear system through a first-order linearization around the observed trajectory:
\begin{equation}
    \delta\mathbf{s}_{t+1}
    =
    \mathbf{A}_t\delta\mathbf{s}_t+\mathbf{w}_t,
    \qquad
    \delta\mathbf{y}_t
    =
    \mathbf{H}_t\delta\mathbf{s}_t+\mathbf{v}_t.
\end{equation}
For analytical clarity, we consider a short local window in which the Jacobians vary slowly and can be approximated as
$\mathbf{A}_t \approx \mathbf{A}$
and
$\mathbf{H}_t \approx \mathbf{H}$.
This local time-invariant approximation is used only for the following analysis. The complete nonlinear-to-linear formulation and assumptions are provided in Appendix~\ref{app:formulation}.

\begin{proposition}[Existence of a same-frame shortcut under spatial redundancy]
\label{prop:spatial_shortcut}
Let the latent observation at time $t$ be partitioned into visible and masked components,
\begin{equation}
    \delta\mathbf{y}_{t}^{V}
    =
    \mathbf{H}_{V}\delta\mathbf{s}_{t}
    +
    \mathbf{v}_{t}^{V},
    \qquad
    \delta\mathbf{y}_{t}^{M}
    =
    \mathbf{H}_{M}\delta\mathbf{s}_{t}
    +
    \mathbf{v}_{t}^{M}.
\end{equation}
If there exists a mapping $\mathbf{C}$ such that
\begin{equation}
\left\lVert
        \mathbf{H}_{M}
        -
        \mathbf{C}\mathbf{H}_{V}
    \right\rVert_{2}
    \leq
\epsilon_s,
\end{equation}
then the same-frame predictor
$
\widehat{\delta\mathbf{y}}_{t}^{M}
=
\mathbf{C}\delta\mathbf{y}_{t}^{V}
$
achieves
\begin{equation}
    \left\lVert
        \delta\mathbf{y}_{t}^{M}
        -\mathbf{C}\delta\mathbf{y}_{t}^{V}
    \right\rVert
    \leq
    \epsilon_s
    \left\lVert
        \delta\mathbf{s}_{t}
    \right\rVert
    +
    \left\lVert
        \mathbf{v}_{t}^{M}
        -
        \mathbf{C}\mathbf{v}_{t}^{V}
    \right\rVert.
\end{equation}
\end{proposition}

Proposition~\ref{prop:spatial_shortcut} provides a sufficient condition under which masked latent regions can be predicted from visible regions at the same temporal instant. Importantly, neither the predictor nor its error bound depends on the temporal transition operator $\mathbf{A}$. Therefore, when strong spatial redundancy exists, low patch-prediction error can be achieved without explicitly exploiting temporal transition dynamics. We refer to this possible solution as a \emph{same-frame spatial shortcut}. The proposition does not imply that every patch-level prediction problem admits such a shortcut, rather, it characterizes when such a solution can exist. The proof is provided in Appendix~\ref{app:proof_spatial_shortcut}.

\begin{proposition}[Temporal recoverability under complete temporal masking]
\label{prop:temporal_recovery}
Suppose the complete latent observation at time $t$ is unavailable. For the preceding $K$ observations, define
\begin{equation}
\mathbf{Y}_{K}
=
\begin{bmatrix}
\delta\mathbf{y}_{t-K}\\[-1mm]
\vdots\\[-1mm]
\delta\mathbf{y}_{t-1}
\end{bmatrix},
\qquad
\mathbf{O}_{K}
=
\begin{bmatrix}
\mathbf{H}\\[-1mm]
\mathbf{H}\mathbf{A}\\[-1mm]
\vdots\\[-1mm]
\mathbf{H}\mathbf{A}^{K-1}
\end{bmatrix},
\qquad
\mathbf{Y}_{K}
=
\mathbf{O}_{K}\delta\mathbf{s}_{t-K}
+
\boldsymbol{\eta},
\end{equation}
where $\boldsymbol{\eta}$ collects process, observation, and local-linearization residuals. If the local system is observable,
$\operatorname{rank}(\mathbf{O}_{K})=d_s$, and
\begin{equation}
\delta\mathbf{y}_{t}
=
\mathbf{H}\mathbf{A}^{K}\delta\mathbf{s}_{t-K}
+
\mathbf{r}_{t},
\qquad
\lVert\mathbf{r}_{t}\rVert
\leq
\epsilon_{\mathrm{dyn}},
\end{equation}
then the predictor
$
\widehat{\delta\mathbf{y}}_{t}
=
\mathbf{H}\mathbf{A}^{K}
\mathbf{O}_{K}^{\dagger}\mathbf{Y}_{K}
$
satisfies
\begin{equation}
\left\lVert
\widehat{\delta\mathbf{y}}_{t}
-
\delta\mathbf{y}_{t}
\right\rVert
\leq
\frac{
\lVert\mathbf{H}\mathbf{A}^{K}\rVert_{2}
}
{
\sigma_{\min}(\mathbf{O}_{K})
}
\lVert\boldsymbol{\eta}\rVert
+
\epsilon_{\mathrm{dyn}}.
\end{equation}
\end{proposition}

The past-only formulation in Proposition~\ref{prop:temporal_recovery} is used for analytical clarity and the actual predictor can exploit context from both temporal directions. Complete temporal masking removes the direct same-frame shortcut characterized in Proposition~\ref{prop:spatial_shortcut}.
Under local observability, however, the masked observation remains recoverable from temporally distributed context through the local transition dynamics. This is a recoverability result rather than a guarantee that a finite-capacity neural network explicitly reconstructs the physical state. The proof is
provided in Appendix~\ref{app:proof_temporal_recovery}.

\paragraph{Design implication.}
The two propositions motivate complementary masking strategies. Patch-level masking provides dense spatial supervision but may admit same-frame shortcuts under strong spatial redundancy. Temporal-slice masking removes all observations at selected temporal indices, eliminating this direct same-frame evidence and increasing reliance on temporally distributed context. ScaleDrive therefore combines patch-level and temporal-slice masking to jointly strengthen spatial and temporal representation learning.

\subsection{Spatial-Temporal Latent Prediction}

Guided by the above analysis, we instantiate ScaleDrive using a joint-embedding predictive architecture on action-free driving videos.

\paragraph{Input and latent prediction.}
Given an action-free driving video
$\mathbf{V}=\{\mathbf{I}_t\}_{t=1}^{T}\in\mathbb{R}^{T\times H\times W\times 3}$,
we partition it into non-overlapping spatiotemporal patches and project them into
visual tokens
$\mathbf{X}=[\mathbf{x}_1,\ldots,\mathbf{x}_N]\in\mathbb{R}^{N\times D}$,
where $N$ and $D$ denote the number and dimension of tokens, respectively. The temporal and spatial coordinates are encoded using 3D RoPE~\cite{su2024roformer}.

Rather than reconstructing RGB pixels, ScaleDrive predicts latent representations of masked spatiotemporal regions. Given a mask set $\mathcal{M}$, the context encoder $E_{\theta}$ processes only the visible tokens, while a predictor $P_{\phi}$ infers the representations at masked positions. The prediction targets are obtained from
the complete token sequence using a momentum target encoder $E_{\bar{\theta}}$:
\begin{equation}
\begin{aligned}
    \mathbf{Z}^{\mathrm{con}}
    = E_{\theta}(\mathbf{X}_{\setminus\mathcal{M}}), \qquad
    \hat{\mathbf{Z}}_{\mathcal{M}}
    = P_{\phi}(\mathbf{Z}^{\mathrm{con}},\mathcal{M}), \qquad
    \mathbf{Z}^{\mathrm{tar}}
    = E_{\bar{\theta}}(\mathbf{X}).
\end{aligned}
\end{equation}
The target encoder shares the same architecture as the context encoder and is updated by an exponential moving average,
$\bar{\theta}\leftarrow\mu\bar{\theta}+(1-\mu)\theta$,
with gradients stopped.

\paragraph{Patch-level spatial masking.}
Following V-JEPA2, we sample spatially contiguous groups of spatiotemporal tokens, denoted by $\mathcal{M}_{p}\subseteq\{1,\ldots,N\}$, and remove them from the context encoder input. The predictor then infers their target representations
from the remaining visible visual context. This provides dense spatial supervision but, as characterized by Proposition~\ref{prop:spatial_shortcut}, may admit solutions that rely heavily on same-frame spatial redundancy.

\paragraph{Temporal-slice masking.}
To complement spatial masking, we additionally remove complete temporal slices. Let $T'$ denote the temporal resolution after patchification and
$\mathcal{T}_{f}\subseteq\{1,\ldots,T'\}$ the sampled temporal indices. The corresponding temporal mask is
\begin{equation}
    \mathcal{M}_{f}
    =
    \left\{
        i \mid \tau(i)\in\mathcal{T}_{f}
    \right\},
\end{equation}
where $\tau(i)$ maps token $i$ to its post-patchification temporal index.
Because all spatial tokens at the selected temporal indices are removed, the predictor cannot directly exploit visible observations from the same temporal slice and must rely more strongly on temporally distributed context. During pretraining, the two masking strategies are combined as
\begin{equation}
    \mathcal{M}
    =
    \mathcal{M}_{p}\cup\mathcal{M}_{f}.
\end{equation}

\paragraph{Latent prediction objective.}
For each masked position $i\in\mathcal{M}$, the predicted latent
$\hat{\mathbf{z}}_i$ is optimized to match the corresponding target representation $\mathbf{z}^{\mathrm{tar}}_i$. Following V-JEPA2, we minimize an $L_1$ regression loss over masked tokens:
\begin{equation}
    \mathcal{L}_{\mathrm{VLP}}
    =
    \frac{1}{|\mathcal{M}|}
    \sum_{i\in\mathcal{M}}
    \left\lVert
        \hat{\mathbf{z}}_i
        -
        \operatorname{sg}
        \left(
            \mathbf{z}^{\mathrm{tar}}_i
        \right)
    \right\rVert_1,
\end{equation}
where $\operatorname{sg}(\cdot)$ denotes the stop-gradient operator.
The context encoder and predictor are optimized jointly, while the target encoder
is updated only through the momentum update.

ScaleDrive therefore retains the standard joint-embedding prediction objective and momentum-target mechanism, while modifying the masking strategy: patch-level prediction provides dense spatial supervision, whereas complete temporal-slice prediction reduces reliance on same-frame spatial shortcuts and strengthens temporal predictive learning.

\subsection{Scaling and Frozen Representation Transfer}

\paragraph{Fixed-compute data scaling.}
ScaleDrive scales representation learning along the data dimension using raw action-free driving videos. Across different pretraining data scales, we keep the encoder architecture, masking formulation, optimization configuration, and overall training budget fixed, while varying only the size of the sampled driving-video pool. This controlled protocol isolates whether increasing the diversity and coverage of raw driving observations improves the quality of the learned representation under a comparable optimization budget.

\paragraph{Frozen representation transfer.}
After pretraining, we discard the predictor and retain the target encoder, denoted by $E_{\mathrm{vis}}$, as the visual representation model. Given an observed driving video $\mathbf{V}$, the encoder produces a sequence of spatiotemporal latent representations:
\begin{equation}
    \mathbf{Z}
    =
    E_{\mathrm{vis}}(\mathbf{V})
    \in
    \mathbb{R}^{N \times D}.
\end{equation}
The parameters of $E_{\mathrm{vis}}$ remain frozen throughout downstream policy training. This design is deliberate: if the visual encoder is jointly fine-tuned with the planner, improvements in downstream planning performance could arise from representation adaptation during policy training. Freezing the encoder makes the downstream performance a more direct measure of the information already learned during action-free video pretraining.

\paragraph{Planning interface.}
Given the frozen video representation $\mathbf{Z}$ and the current ego status $\mathbf{e}$, the downstream planner models
\begin{equation}
    p_{\omega}
    \left(
        \mathbf{Y}
        \mid
        \mathbf{Z},
        \mathbf{e}
    \right),
\end{equation}
where $\mathbf{Y}$ denotes the future ego trajectory and $\omega$ contains only the trainable parameters of the planning module. With a set of learnable ego queries
$\mathbf{Q}_{e}\in\mathbb{R}^{M\times D}$,
the trajectory decoder predicts
\begin{equation}
    \hat{\mathbf{Y}}
    =
    D_{\omega}
    \left(
        \mathbf{Q}_{e},
        \mathbf{Z},
        \mathbf{e}
    \right).
\end{equation}
The downstream decoder follows established end-to-end planning objectives~\citep{yao2026drivesuprim,kirby2026driving,wang2026chainflow}. The architectural and optimization details are provided in Appendix~\ref{sec:planning}. This deliberately simple interface reduces architectural confounding and allows downstream performance to more directly reflect the quality of the pretrained driving representation.

\paragraph{Heterogeneous planner integration.}
ScaleDrive is not tied to the primary trajectory decoder. Its frozen visual encoder can serve as a plug-in module and be integrated into heterogeneous planners through simple planner-specific interfaces. BEV-based policies fuse ScaleDrive representation with the native BEV features, whereas camera-view-based policies replace the native visual representation with ScaleDrive. The detailed integration strategies are provided in Appendix~\ref{app:cross_planner_configs}.

\section{Experiment}
\label{Experiment}

\subsection{Experimental Setup}

We pretrain ScaleDrive with OpenDV-YouTube~\citep{yang2024generalized}, a large-scale collection of internet driving videos without action or task-specific annotations. The raw driving videos are segmented into 4-second clips sampled at 2\,Hz, yielding a total of 6M video clips for video latent prediction pretraining. We use NAVSIMv1~\citep{dauner2024navsim} and NAVSIMv2~\citep{navsimv2} as the primary end-to-end planning benchmarks. Beyond the main planning evaluation, we assess the ScaleDrive representation quality through ego-motion and monocular depth probing, evaluate cross-policy transfer with different planner architectures, and study cross-benchmark generalization on HUGSIM~\citep{zhou2025hugsim} and Bench2Drive~\citep{jia2024bench2drive}. Details on the training configurations and evaluation protocols are provided in Appendix~\ref{sec:implementation_details}.

\begin{table}[t]
\centering
\caption{
\textbf{Planning performance on NAVSIMv1.}~``Enc. Frozen'' indicates whether the visual encoder is kept frozen during
downstream planner training. $\dagger$ denotes training without perception-task labels.
}
\setlength{\tabcolsep}{8pt}
\resizebox{\textwidth}{!}{
\begin{tabular}{l|c|c|cc|ccc|c}
\toprule
Method & Inputs & Enc. Frozen &  NC & DAC &
EP & TTC & Comf. & PDMS \\
\midrule
Transfuser~\cite{chitta2022transfuser} & L\&C & \xmark & 97.7 & 92.8 & 79.2 & 92.8 & 100.0 & 84.0 \\
HydraMDP~\cite{li2024hydra} & L\&C & \xmark & 98.3 & 96.0 & 78.7 & 94.6 & 100.0 & 86.5 \\
DiffusionDrive~\cite{liao2025diffusiondrive} & L\&C & \xmark & 98.2 & 96.2 & 82.2 & 94.7 & 100.0 & 88.1 \\
World4Drive~$\dagger$~\cite{zheng2025world4drive} & C & \xmark & 97.4 & 94.3 & 79.9 & 92.8 & 100.0 & 85.1 \\
Drive-JEPA~$\dagger$~\cite{drivejepa} & C & \xmark & 98.7 & 96.2 & 82.9 & 95.5 & 100.0 & 89.0 \\ 
ScaleDrive$\dagger$ & C & \cmark & 99.0 & 96.3 & 83.0 & 96.1 & 100.0 & 89.1 \\
\midrule

ReCogDrive~\cite{li2026recogdrive} & C & \cmark & 97.9 & 97.3 & 87.3 & 94.9 & 100.0 & 90.8 \\
SGDrive~\cite{li2026sgdrive} & C & \cmark & 98.6 & 97.8 & 85.8 & 96.2 & 100.0 & 91.1 \\
iPad~\cite{guo2026ipad} & C &\xmark & 98.6 & 98.3 & 88.0 & 94.9 & ~~99.9 & 91.7 \\
DrivoR~\cite{kirby2026driving} & C & \xmark & 98.9 & 98.3 & 89.1 & 96.2 & 100.0 & 93.1 \\
Drive-JEPA~\cite{drivejepa} & C & \xmark & 99.1 & 98.2 & \textbf{91.4} & 96.2 & 100.0 & 93.7 \\
ChainFlow-VLA~\cite{wang2026chainflow} & C & \xmark & 98.8 & \textbf{98.6} & 90.8 & 96.1 & 100.0 & 93.6 \\
\midrule

\rowcolor{lightblue}
\textbf{ScaleDrive} & C  & \cmark & \textbf{99.3} & 98.4 & 90.6 & \textbf{97.0} & 100.0 & \textbf{93.8} \\

\bottomrule

\end{tabular}
}
\label{tab:navsim1}
\end{table}

\begin{table}[t]
  \centering
  \caption{
  \textbf{Planning performance on NAVSIMv2.}~ScaleDrive achieves strong overall performance under the more comprehensive closed-loop evaluation metrics.
  }
  \label{navsimv2}
  \setlength{\tabcolsep}{8pt}
  \resizebox{\linewidth}{!}{
  \begin{tabular}{l|cccc|ccccc|c}
  \toprule
  Method & NC & DAC & DDC & TLC & EP & TTC & LK & HC & EC & EPDMS \\
  \midrule
  DiffusionDrive~\cite{liao2025diffusiondrive}
  & 98.2 & 95.9 & 99.4 & 99.8 
  & 87.5 & 97.3 & 96.8 & \textbf{98.3} & 87.7 & 84.5 \\

  DiffusionDriveV2~\cite{zou2025diffusiondrivev2}
  & 97.7 & 96.6 & 99.2 & 99.8 
  & 88.9 & 97.2 & 96.0 & 97.8 & \textbf{91.0} & 87.5 \\

  SparseDriveV2~\cite{sun2026sparsedrivev2}
  & 98.1 & 98.1 & 99.6 & 99.8 
  & \textbf{91.1} & 97.3 & 96.9 & 98.2 & 78.4 & 90.1 \\
  
  Latent-WAM~\cite{wang2026latent}
  & 98.1 & 97.3 & 99.6 & 99.8
  & 87.7 & 97.3 & \textbf{97.6} & 98.1 & 87.3 & 89.3 \\

  \midrule
  \rowcolor{lightblue}
  \textbf{ScaleDrive} & \textbf{99.3} & \textbf{98.9} & \textbf{99.7} & 99.6
  & 85.4 & \textbf{98.9} & 95.9 & 98.2 & 82.1 & \textbf{90.5} \\ 
  \bottomrule
  \end{tabular}}
\end{table}

\subsection{Main Planning Results}

Tables~\ref{tab:navsim1} and~\ref{navsimv2} report the planning performance of ScaleDrive on NAVSIMv1 and NAVSIMv2.  ScaleDrive achieves strong planning performance on both benchmarks, reaching a PDMS of 93.8 on NAVSIMv1 and an EPDMS of 90.5 on NAVSIMv2. The pretrained video encoder remains frozen throughout downstream policy training, indicating that the resulting performance primarily reflects the quality and transferability of the ScaleDrive representation.

\begin{table}[!t]
\centering

\caption{
\textbf{Transfer across end-to-end driving policies}. Downstream training and evaluation settings are kept fixed within each policy.
}
\label{tab:policy_generalization}
\setlength{\tabcolsep}{10pt}
\resizebox{\textwidth}{!}{
\begin{tabular}{l|c|c|cccc|c}
\toprule
Method & Trajectory Modeling & Representation &  NC & DAC & TTC & EP
& PDMS \\
\midrule
LTF &  & BEV &  97.4 & 92.8 & 92.4 & 79.0 & 83.8 \\
LTF(V-JEPA2) &  Regression & w. V-JEPA2 & 97.8 & 91.7 & 92.4 & 78.8 & 83.2{\scriptsize(-0.6)} \\
\rowcolor{lightblue}
LTF(ScaleDrive) &  & w. ScaleDrive & \textbf{98.2} & \textbf{94.0} & \textbf{93.6} & \textbf{81.6} & \textbf{86.2}{\scriptsize(+2.4)} \\
\midrule
DiffusionDrive &  & BEV &  98.1 & 94.8 & 94.1 & 81.1 & 86.6 \\
DiffusionDrive(V-JEPA2) & Diffusion  & w. V-JEPA2  & \textbf{98.6} & 95.1 & 95.1 & 81.8 & 87.4{\scriptsize(+0.8)} \\
\rowcolor{lightblue}
DiffusionDrive(ScaleDrive) &  & w. ScaleDrive & \textbf{98.6} & \textbf{95.8} & \textbf{95.3} & \textbf{82.3} & \textbf{88.1}{\scriptsize(+1.5)} \\
\midrule
DrivoR&  & DINOv2  & 98.9 & \textbf{98.3} & 96.2 & 89.1 & 93.1 \\
DrivoR(V-JEPA2) & Regression & w. V-JEPA2 & 99.1 & 97.9 & 96.5 & 87.9 & 92.3{\scriptsize(-0.8)} \\
\rowcolor{lightblue}
DrivoR(ScaleDrive)&  & w. ScaleDrive & \textbf{99.2} & 98.0  & \textbf{96.7} & \textbf{89.8} & \textbf{93.4}{\scriptsize(+0.3)} \\
\midrule
ChainFlow&  & DINOv2  & 98.9 & 98.1 & \textbf{96.2} & 88.9 & 92.6 \\
ChainFlow(V-JEPA2) & Autoregressive & w. V-JEPA2 & 98.8 & 97.5 & 95.6 & 89.2 & 92.1{\scriptsize(-0.5)} \\
\rowcolor{lightblue}
ChainFlow(ScaleDrive) &  & w. ScaleDrive & \textbf{99.0} & \textbf{98.3} & \textbf{96.2} & \textbf{90.5} & \textbf{93.2}{\scriptsize(+0.6)} \\

\bottomrule
\end{tabular}
}

\end{table}

\begin{figure}[t]
    \centering
    \includegraphics[width=1.0\linewidth]{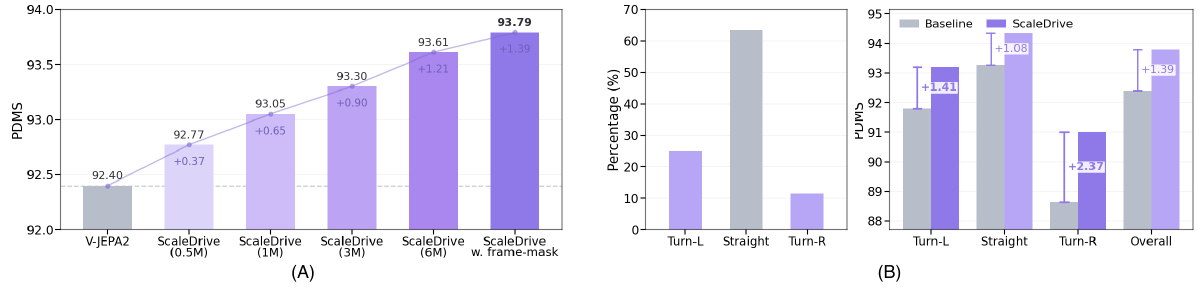}
    \caption{(A)~Data scaling performance of ScaleDrive. (B)~Command-wise gains under the imbalanced training driving action distribution.}
    \label{fig:roadmap_gain}
\end{figure}

We study how ScaleDrive scales with action-free driving videos for downstream planning. Fig.~\ref{fig:roadmap_gain}~(A) illustrates how planning performance scales with the amount of action-free driving-video pretraining. As the number of video clips increases from 0.5M to 6M, PDMS consistently improves from 92.77 to 93.61, compared with 92.40 using the official V-JEPA2 representation. The gain trend empirically shows that increasing the diversity and coverage of driving videos improves the quality of the frozen representation under a fixed training budget. Beyond data scaling, we introduce temporal-slice masking to strengthen temporal modeling during pretraining, following Proposition~\ref{prop:temporal_recovery}. At the same data scale, the additional temporal-slice masking further improves planning performance.

A key motivation for scaling action-free video pretraining is to expose the model to a broader diversity of driving scenes. We investigate whether this broader scene coverage translates into planning gains in long-tail driving scenes. We analyze performance across different driving commands in Fig.~\ref{fig:roadmap_gain}~(B). The downstream training distribution is highly imbalanced, with turning commands following a clear long-tail distribution. ScaleDrive improves planning performance across all commands, with more pronounced gains on long-tail turning maneuvers. These results suggest that the broader scene coverage provided by large-scale video pretraining is particularly beneficial for long-tail driving situations that are less frequently observed during downstream policy training.

\subsection{Does ScaleDrive Transfer Across Driving Policies?}
A transferable driving representation should provide consistent benefits across different policy designs. We evaluate the frozen ScaleDrive representation with four end-to-end driving policies while keeping the training settings fixed within each policy. These policies span regression, diffusion, and autoregressive trajectory modeling, with native visual representations ranging from BEV~\citep{chitta2022transfuser} to DINOv2~\citep{oquab2023dinov2}. As shown in Table~\ref{tab:policy_generalization}, ScaleDrive consistently improves planning performance across all architectures. In contrast, V-JEPA2 does not yield similarly consistent improvements. This comparison indicates that scaling action-free pretraining on driving videos yields representations that transfer across heterogeneous downstream policies.

\begin{table}[t]
\centering
\caption{
\textbf{Cross-benchmark generalization on HUGSIM.}
}
\label{tab:cross_scenario_generalization}
\setlength{\tabcolsep}{5pt}
\resizebox{\linewidth}{!}{
\begin{tabular}{l|c|ccccc|ccccc}
\toprule
& & \multicolumn{5}{c|}{Route Completion $\uparrow$}
& \multicolumn{5}{c}{HD-Score $\uparrow$} \\
Planner & Representation
& Easy & Medium & Hard & Extreme & Avg.
& Easy & Medium & Hard & Extreme & Avg. \\
\midrule

LTF
& BEV
& 0.684 & 0.407 & \textbf{0.369} & 0.255 & 0.410
& 0.528 & 0.246 & \textbf{0.198} & 0.081 & 0.246 \\

LTF
& w. V-JEPA2
& 0.738 & 0.400 & 0.298 & \textbf{0.279} & 0.412
& 0.630 & 0.228 & 0.141 & 0.113 & 0.251 \\

\rowcolor{lightblue}
LTF
& w. ScaleDrive
& \textbf{0.742} & \textbf{0.496} & 0.329 & 0.259 & \textbf{0.448}
& \textbf{0.638} & \textbf{0.361} & 0.180 & \textbf{0.119} & \textbf{0.315} \\

\midrule

ChainFlow
& DINOv2
& 0.834 & 0.433 & 0.258 & 0.273 & 0.430
& \textbf{0.705} & 0.313 & 0.116 & 0.122 & 0.296 \\

ChainFlow
& w. V-JEPA2
& 0.782 & 0.443 & 0.315 & 0.252 & 0.432
& 0.640 & 0.342 & 0.168 & 0.108 & 0.302 \\

\rowcolor{lightblue}
ChainFlow
& w. ScaleDrive
& \textbf{0.843} & \textbf{0.561} & \textbf{0.362} & \textbf{0.274} & \textbf{0.500}
& 0.693 & \textbf{0.442} & \textbf{0.210} & \textbf{0.123} & \textbf{0.361} \\

\bottomrule
\end{tabular}}
\end{table}

\begin{table}[!t]
\centering

\caption{
\textbf{Cross-benchmark generalization on Bench2Drive.}}
\label{tab:bench2drive}
\setlength{\tabcolsep}{12pt}
\resizebox{0.9\linewidth}{!}{
\begin{tabular}{l|c|cccc}
\toprule
Planner & Representation & Driving Score$\uparrow$ & Success Rate$\uparrow$ & Efficiency$\uparrow$ & Comfort$\uparrow$ \\
\midrule
TFv6 & BEV & 95.20 & 86.80 & - & - \\
TFv6 & w. V-JEPA2 & 95.00 & 85.91 & 202.33 & 15.45 \\
\rowcolor{lightblue}TFv6 & w. ScaleDrive & \textbf{96.65}({+1.45}) & \textbf{90.45}(+3.65) & \textbf{210.22} & \textbf{18.03}\\
\bottomrule
\end{tabular}}
\end{table}

\subsection{Does ScaleDrive Transfer to Different Driving Scenarios?}

A central motivation of ScaleDrive is to investigate whether action-free driving videos can induce transferable driving representations that generalize beyond the downstream training distribution. We evaluate ScaleDrive on HUGSIM~\citep{zhou2025hugsim} and Bench2Drive~\citep{jia2024bench2drive}, two closed-loop benchmarks with distinct scenario distributions and evaluation protocols. Table~\ref{tab:cross_scenario_generalization} evaluates cross-benchmark generalization on HUGSIM. ScaleDrive improves the average Route Completion and HD-Score with different planner architectures and original representations, with clear benefits in the Medium and Hard scenarios. Table~\ref{tab:bench2drive} further shows consistent improvements over the strong TFv6~\citep{nguyen2026lead} planner across closed-loop metrics on Bench2Drive. These results indicate that the learned representation generalizes across distinct driving environments, scenario distributions, and evaluation protocols.

\begin{figure}[!t]
      \centering
      \includegraphics[width=1.0\linewidth]{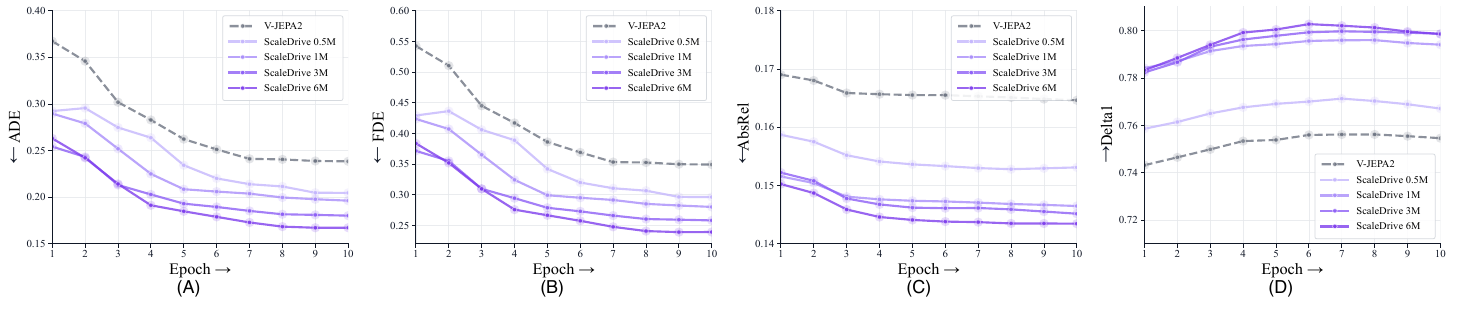}
      \caption{
      \textbf{Video representation probing.}~(A--B) Ego-motion estimation with ADE~$\downarrow$ and FDE~$\downarrow$;  (C--D) monocular depth estimation with AbsRel~$\downarrow$ and $\delta_1$~$\uparrow$.
      }
      \label{fig:motion_probe}
  \end{figure}

\subsection{Does Action-Free Video Latent Prediction Improve Driving Representations?}

Proposition~\ref{prop:temporal_recovery} suggests that temporal-slice prediction places stronger requirements on representing temporally evolving scene states. We therefore directly probe whether motion information becomes more accessible from the frozen representation through ego-motion estimation. As a complementary assessment of spatial representation quality, we further use monocular depth probing on KITTI~\citep{kitti} to assess whether scaling action-free pretraining also strengthens spatial-geometric information. In both settings, the pretrained video encoder remains frozen and only the task-specific probe is optimized. Detailed architectures and optimization settings are provided in the Appendix~\ref{app:frozen-probes}. As shown in Fig.~\ref{fig:motion_probe}, performance consistently improves as the amount of driving-video pretraining increases for both ego-motion and depth estimation. This trend indicates that scaling video latent prediction progressively strengthens the temporal and geometric information encoded in the frozen representation.

\begin{figure}[t]
      \centering
      \includegraphics[width=1.0\linewidth]{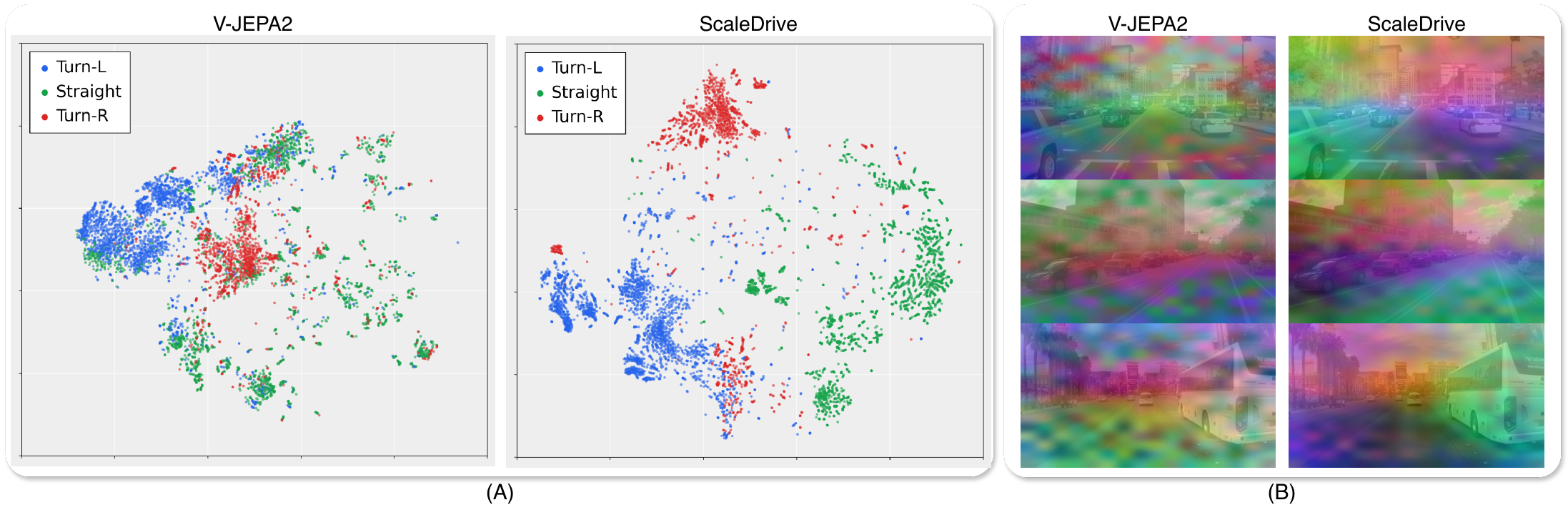}
      \caption{
      \textbf{Qualitative result of representations.}~(A) t-SNE visualization. (B) PCA visualization of dense video features across representative driving scenes.
      }
      \label{fig:tsne_pca}
\end{figure}

\begin{figure}[!t]
      \centering
      \includegraphics[width=1.0\linewidth]{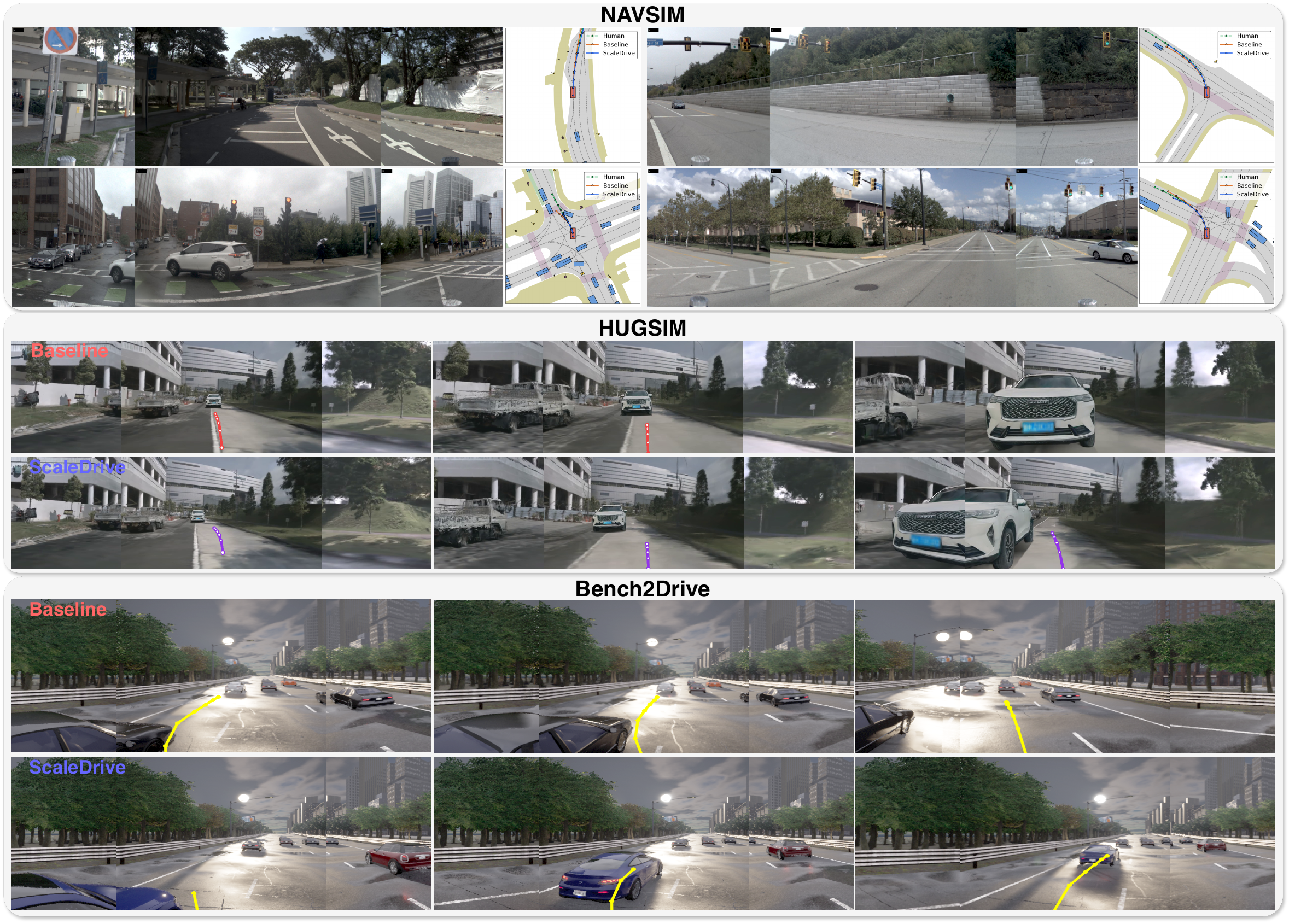}
      \caption{
      \textbf{Planning results on NAVSIM, HUGSIM and Bench2Drive.}
      }
      \label{fig:quant}
\end{figure}

\subsection{Qualitative results}

\paragraph{Visual representation.}~Fig.~\ref{fig:tsne_pca} provides qualitative comparisons of the spatiotemporal representations learned by V-JEPA2 and ScaleDrive. In the t-SNE visualizations shown in Fig.~\ref{fig:tsne_pca}~(A), ScaleDrive exhibits more clearly separated feature distributions across different driving behaviors. The PCA visualizations in Fig.~\ref{fig:tsne_pca}~(B) further show more organized spatial feature patterns across representative driving scenes. These observations suggest that scaling action-free video latent prediction pretraining yields more structured representations.

\paragraph{End-to-end Planning.}~Fig.~\ref{fig:quant} shows representative planning cases on NAVSIM, HUGSIM, and Bench2Drive. On NAVSIM, the predicted trajectories more closely follow the reference expert behavior across diverse road geometries and intersection scenarios. On HUGSIM and Bench2Drive, ScaleDrive exhibits safer closed-loop behavior, including appropriate responses to approaching vehicles and potentially unsafe interactions. These examples illustrate that stronger spatiotemporal representations can translate into more reliable downstream planning across driving environments.

\section{Conclusion}
We present ScaleDrive, a scalable representation-learning framework that learns driving representations from action-free videos through video latent prediction. By scaling pretraining data while keeping the downstream encoder frozen, we show that larger collections of driving videos consistently improve representation quality and translate into stronger planning performance. The learned representations capture increasingly useful motion and geometric cues, and generalize across different planning architectures and driving benchmarks. These results suggest that action-free video pretraining provides a promising direction for scaling autonomous driving beyond the availability of action-annotated data.

\clearpage
\bibliography{biblio}
\bibliographystyle{template_conference}

\clearpage
\appendix
\appendix

\section{Theoretical Analysis}
\label{app:theory}

\subsection{Dynamic-System Formulation and Assumptions}
\label{app:formulation}

The main-text analysis abstracts a driving video as a sequence of observations generated by an underlying latent dynamic system,
\begin{equation}
    \mathbf{s}_{t+1}=F(\mathbf{s}_t)+\mathbf{w}_t,\qquad \mathbf{y}_t=G(\mathbf{s}_t)+\mathbf{v}_t,
\end{equation}
where $\mathbf{s}_t\in\mathbb{R}^{d_s}$ denotes the latent physical state of the driving scene, including ego motion, surrounding-agent states, and local scene geometry and $\mathbf{y}_t$ denotes a temporally indexed latent observation corresponding to the predictive-learning target, and $\mathbf{w}_t$ and $\mathbf{v}_t$ denote process and observation noise, respectively. This formulation is an analytical abstraction rather than an exact model of the neural architecture: in particular, we do not assume that the learned representation explicitly corresponds to $\mathbf{s}_t$ or that the Transformer target representation depends exclusively on the physical state at a single instant.

For a state perturbation $\delta\mathbf{s}_t$ around the observed trajectory, a first-order Taylor expansion gives
\begin{equation}
    F(\mathbf{s}_t+\delta\mathbf{s}_t)\approx F(\mathbf{s}_t)+\left.\frac{\partial F}{\partial\mathbf{s}}\right|_{\mathbf{s}_t}\delta\mathbf{s}_t,\qquad G(\mathbf{s}_t+\delta\mathbf{s}_t)\approx G(\mathbf{s}_t)+\left.\frac{\partial G}{\partial\mathbf{s}}\right|_{\mathbf{s}_t}\delta\mathbf{s}_t.
\end{equation}
Defining
\begin{equation}
    \mathbf{A}_t=\left.\frac{\partial F}{\partial\mathbf{s}}\right|_{\mathbf{s}_t},\qquad \mathbf{H}_t=\left.\frac{\partial G}{\partial\mathbf{s}}\right|_{\mathbf{s}_t},
\end{equation}
yields the local linear model used in the main text,
\begin{equation}
    \delta\mathbf{s}_{t+1}=\mathbf{A}_t\delta\mathbf{s}_t+\mathbf{w}_t,\qquad \delta\mathbf{y}_t=\mathbf{H}_t\delta\mathbf{s}_t+\mathbf{v}_t.
\end{equation}

The analysis relies on the following local assumptions.

\paragraph{Assumption A1 (Local smoothness).} $F$ and $G$ are sufficiently smooth in a neighborhood of the observed trajectory over a short video clip, such that the first-order Taylor approximation is locally valid. Higher-order approximation residuals are absorbed into the disturbance terms used in the subsequent analysis.

\paragraph{Assumption A2 (Local time invariance).} Over the short temporal window considered in Proposition~\ref{prop:temporal_recovery}, the Jacobians vary sufficiently slowly that $\mathbf{A}_t\approx\mathbf{A}$ and $\mathbf{H}_t\approx\mathbf{H}$. This assumption is introduced only to obtain a compact local observability analysis using powers of $\mathbf{A}$ and it does not imply that real driving dynamics are globally linear or time invariant.

\paragraph{Assumption A3 (Local spatial redundancy).} Proposition~\ref{prop:spatial_shortcut} assumes the existence of a mapping $\mathbf{C}$ such that $\lVert\mathbf{H}_{M}-\mathbf{C}\mathbf{H}_{V}\rVert_2\leq\epsilon_s$ for the analyzed frame. This is a sufficient local condition under which the masked component can be approximately inferred from the visible component of the same temporal observation. It does not assume that every masked region in every driving scene can be reconstructed from same-frame information.

\paragraph{Assumption A4 (Local temporal observability).} For the local linear system in Proposition~\ref{prop:temporal_recovery}, the observability matrix satisfies $\operatorname{rank}(\mathbf{O}_K)=d_s$ over the analyzed temporal window. This does not require global observability of all driving scenes and it only states that the state directions relevant to the local analysis are sufficiently constrained by the available temporal observations.

\subsection{Proof of Proposition~\ref{prop:spatial_shortcut}}
\label{app:proof_spatial_shortcut}

Let the visible and masked components of the latent observation at time $t$ be
\begin{equation}
    \delta\mathbf{y}_t^{V}=\mathbf{H}_{V}\delta\mathbf{s}_t+\mathbf{v}_t^{V},\qquad \delta\mathbf{y}_t^{M}=\mathbf{H}_{M}\delta\mathbf{s}_t+\mathbf{v}_t^{M}.
\end{equation}

Consider the same-frame predictor $\widehat{\delta\mathbf{y}}_t^{M}=\mathbf{C}\delta\mathbf{y}_t^{V}$. The prediction residual can be written as
\begin{equation}
    \delta\mathbf{y}_t^{M}-\mathbf{C}\delta\mathbf{y}_t^{V}=\left(\mathbf{H}_{M}-\mathbf{C}\mathbf{H}_{V}\right)\delta\mathbf{s}_t+\left(\mathbf{v}_t^{M}-\mathbf{C}\mathbf{v}_t^{V}\right).
\end{equation}

Applying the triangle inequality and the induced matrix norm gives
\begin{equation}
    \left\lVert\delta\mathbf{y}_t^{M}-\mathbf{C}\delta\mathbf{y}_t^{V}\right\rVert\leq \left\lVert\mathbf{H}_{M}-\mathbf{C}\mathbf{H}_{V}\right\rVert_2\left\lVert\delta\mathbf{s}_t\right\rVert+\left\lVert\mathbf{v}_t^{M}-\mathbf{C}\mathbf{v}_t^{V}\right\rVert.
\end{equation}

Under Assumption~A3, $\lVert\mathbf{H}_{M}-\mathbf{C}\mathbf{H}_{V}\rVert_2\leq\epsilon_s$, and therefore
\begin{equation}
    \left\lVert\delta\mathbf{y}_t^{M}-\mathbf{C}\delta\mathbf{y}_t^{V}\right\rVert\leq \epsilon_s\left\lVert\delta\mathbf{s}_t\right\rVert+\left\lVert\mathbf{v}_t^{M}-\mathbf{C}\mathbf{v}_t^{V}\right\rVert,
\end{equation}
which is the bound stated in Proposition~\ref{prop:spatial_shortcut}.

Crucially, neither the same-frame predictor nor its error bound contains the temporal transition operator $\mathbf{A}$. Therefore, whenever the spatial-redundancy condition in Assumption~A3 approximately holds, a low masked-prediction error can in principle be achieved without explicitly exploiting temporal transition dynamics. Proposition~\ref{prop:spatial_shortcut} should therefore be interpreted as establishing the existence of a possible same-frame shortcut under spatial redundancy, rather than claiming that all patch-level prediction is solved through such a shortcut.

\subsection{Proof of Proposition~\ref{prop:temporal_recovery}}
\label{app:proof_temporal_recovery}

Under Assumptions~A1 and A2, consider the local linear dynamics over the temporal window $[t-K,t]$,
\begin{equation}
    \delta\mathbf{s}_{\ell+1}=\mathbf{A}\delta\mathbf{s}_{\ell}+\mathbf{w}_{\ell},\qquad \delta\mathbf{y}_{\ell}=\mathbf{H}\delta\mathbf{s}_{\ell}+\mathbf{v}_{\ell}.
\end{equation}

Starting from $\delta\mathbf{s}_{t-K}$, recursively applying the transition dynamics gives, for $j=0,\ldots,K-1$,
\begin{equation}
    \delta\mathbf{s}_{t-K+j}=\mathbf{A}^{j}\delta\mathbf{s}_{t-K}+\boldsymbol{\xi}_{j},
\end{equation}
where $\boldsymbol{\xi}_{j}$ aggregates the accumulated process disturbances and local approximation residuals up to the $j$-th step. The corresponding observations satisfy
\begin{equation}
    \delta\mathbf{y}_{t-K+j}=\mathbf{H}\mathbf{A}^{j}\delta\mathbf{s}_{t-K}+\mathbf{H}\boldsymbol{\xi}_{j}+\mathbf{v}_{t-K+j}.
\end{equation}

Stacking the $K$ preceding observations yields
\begin{equation}
    \mathbf{Y}_{K}=
    \begin{bmatrix}
        \delta\mathbf{y}_{t-K}\\
        \delta\mathbf{y}_{t-K+1}\\
        \vdots\\
        \delta\mathbf{y}_{t-1}
    \end{bmatrix}
    =
    \mathbf{O}_{K}\delta\mathbf{s}_{t-K}+\boldsymbol{\eta},
\end{equation}
where
\begin{equation}
    \mathbf{O}_{K}=
    \begin{bmatrix}
        \mathbf{H}\\
        \mathbf{H}\mathbf{A}\\
        \vdots\\
        \mathbf{H}\mathbf{A}^{K-1}
    \end{bmatrix},
\end{equation}
and $\boldsymbol{\eta}$ aggregates process disturbances, observation noise, and local-linearization residuals.

Under Assumption~A4, $\operatorname{rank}(\mathbf{O}_{K})=d_s$, so $\mathbf{O}_{K}$ has full column rank and $\mathbf{O}_{K}^{\dagger}\mathbf{O}_{K}=\mathbf{I}$. The least-squares estimate of the initial state in the local window is therefore
\begin{equation}
    \hat{\mathbf{s}}_{t-K}=\mathbf{O}_{K}^{\dagger}\mathbf{Y}_{K}.
\end{equation}

Substituting the stacked observation model gives
\begin{equation}
    \hat{\mathbf{s}}_{t-K}-\delta\mathbf{s}_{t-K}=\mathbf{O}_{K}^{\dagger}\boldsymbol{\eta}.
\end{equation}

Taking norms and using $\lVert\mathbf{O}_{K}^{\dagger}\rVert_2=1/\sigma_{\min}(\mathbf{O}_{K})$ yields
\begin{equation}
    \left\lVert\hat{\mathbf{s}}_{t-K}-\delta\mathbf{s}_{t-K}\right\rVert\leq\frac{\lVert\boldsymbol{\eta}\rVert}{\sigma_{\min}(\mathbf{O}_{K})}.
\end{equation}

For the completely masked target instant $t$, write the local target observation as
\begin{equation}
    \delta\mathbf{y}_{t}=\mathbf{H}\mathbf{A}^{K}\delta\mathbf{s}_{t-K}+\mathbf{r}_{t},
\end{equation}
where $\mathbf{r}_{t}$ collects accumulated process disturbances,
target-observation noise, and local-linearization residuals, and satisfies
$\lVert\mathbf{r}_{t}\rVert\leq\epsilon_{\mathrm{dyn}}$.

Using the recovered state, define
\begin{equation}
    \widehat{\delta\mathbf{y}}_{t}=\mathbf{H}\mathbf{A}^{K}\hat{\mathbf{s}}_{t-K}.
\end{equation}

The prediction residual is then
\begin{equation}
    \widehat{\delta\mathbf{y}}_{t}-\delta\mathbf{y}_{t}=\mathbf{H}\mathbf{A}^{K}\left(\hat{\mathbf{s}}_{t-K}-\delta\mathbf{s}_{t-K}\right)-\mathbf{r}_{t}.
\end{equation}

Applying the triangle inequality and the state-estimation bound gives
\begin{equation}
    \left\lVert\widehat{\delta\mathbf{y}}_{t}-\delta\mathbf{y}_{t}\right\rVert\leq\frac{\lVert\mathbf{H}\mathbf{A}^{K}\rVert_2}{\sigma_{\min}(\mathbf{O}_{K})}\lVert\boldsymbol{\eta}\rVert+\epsilon_{\mathrm{dyn}},
\end{equation}
which is the bound stated in Proposition~\ref{prop:temporal_recovery}.

This result establishes a recoverability statement rather than a learning guarantee. Once the complete observation at the target temporal index is removed, the direct same-frame solution characterized by Proposition~\ref{prop:spatial_shortcut} is unavailable. Under local observability, the missing observation can nevertheless be inferred from information distributed across preceding temporal observations through the local transition dynamics. The analysis therefore motivates temporal-slice masking rather than guaranteeing that a finite-capacity neural network recovers the complete physical state.

\section{Extended Related Works}
\label{app:related}

\subsection{Vision Representation for Autonomous Driving}
Visual representation learning is a central component of end-to-end autonomous driving~\citep{chen2024end, chib2023recent, ma2024vision}. Early approaches typically construct task-specific intermediate representations, such as bird's-eye-view (BEV) features~\citep{bevformer, bevdet, fiptr,yang2023bevformer, liao2025diffusiondrive, gui2025trajdiff} for spatially aligned scene understanding and occupancy representations~\citep{occ3d, occformer, wei2023surroundocc} for modeling dense 3D structure and free-space geometry. More recently, the emergence of large-scale pretrained vision and video foundation models~\citep{oquab2023dinov2, tschannen2025siglip, simeoni2025dinov3, wan, cogvideox} and vision-language models~\citep{wang2024qwen2, chen2024internvl, bai2025qwen3} have motivated the use of generic foundation-model features as stronger visual priors for downstream driving~\citep{kirby2026driving, gui2026bridging, yan2026causaldrive, zhang2025epona, li2026recogdrive, yan2026ad, jiang2026senna, gui2026explicit}. Along this direction, video-based joint-embedding predictive architecture~\citep{bardes2024revisiting, vjepa2} has recently been adapted to autonomous driving. Drive-JEPA~\citep{drivejepa} uses self-supervised pretraining on driving videos to learn planning-oriented visual representations, while Auto-JEPA~\citep{yang2026auto} extends joint-embedding prediction toward action-oriented future-intent modeling. However, it remains underexplored whether video representation learning with only action-free driving video itself can be systematically scaled and transferred to end-to-end planning.

\subsection{Scaling in Autonomous Driving}
Recent works have investigated how autonomous driving policies scale with increasing amounts of action-supervised driving data~\citep{naumann2025data, zheng2024data}, showing that larger training sets can yield consistent gains in end-to-end driving performance. Subsequent studies further improve this scaling behavior through large-scale simulation and world-model supervision~\citep{tian2026simscale,li2026drivevla}, but the resulting policies still rely on action supervision during training. In parallel, several works explore scaling representation learning for autonomous driving~\citep{kong2025largead,li2026stellar}, demonstrating the benefits of larger-scale data and models for perception tasks. However, whether action-free driving representations exhibit similar scaling behavior, and whether such gains transfer to downstream planning, remains underexplored. We address this gap by scaling video latent prediction on large-scale action-free driving videos and evaluating the resulting frozen representations across diverse planning architectures and benchmarks.

\section{Implementation Details}
\label{sec:implementation_details}

\subsection{ScaleDrive Video latent prediction pretraining.}
\label{sec:pretrain}

\begin{table}[t]
\centering
\caption{
\textbf{ScaleDrive video latent prediction pretraining configuration.}
}
\label{tab:scaledrive_video_latent_training}
\setlength{\tabcolsep}{6pt}
\begin{tabular}{lll}
\toprule
\textbf{Category} & \textbf{Hyperparameter} & \textbf{Value} \\
\midrule

\multirow{7}{*}{Model}
& Backbone & ViT-Large with 3D RoPE \\
& Resolution & $256 \times 256$ \\
& Frames per clip & 8 \\
& Patch size & $16 \times 16$ \\
& Tubelet size & 2 \\
& Predictor dimension & 256 \\
& Downstream checkpoint & EMA target encoder \\

\midrule

\multirow{4}{*}{Masking}
& Masking strategy & Multi-group patch + temporal-slice \\
& Small-patch mask & 8 groups, spatial scale 0.15 \\
& Large-patch mask & 2 groups, spatial scale 0.70 \\
& Temporal-slice mask & 1 frame \\

\midrule

\multirow{9}{*}{Training}
& Trainable parameters & 304.148M \\
& Optimizer & AdamW \\
& Start learning rate & $5.25 \times 10^{-4}$ \\
& Final learning rate & $1 \times 10^{-6}$ \\
& Number of GPUs & 16 $\times$ NVIDIA A800 \\
& Global batch size & 4096 \\
& Training epochs & 20 \\
& Iterations per epoch & 300 \\
& EMA momentum & 0.99925 \\

\midrule

\multirow{2}{*}{Data}
& Dataset & OpenDV-YouTube \\
& Dataset size & 0.5M / 1M / 3M / 6M \\

\bottomrule
\end{tabular}
\end{table}

ScaleDrive follows the general video latent prediction framework of V-JEPA2. The encoder adopts a ViT-Large architecture with 3D RoPE and is initialized from the official V-JEPA2 ViT-L checkpoint.  Each input consists of eight frames sampled at 2\,Hz and resized to $256\times256$ pixels, with a spatial patch size of $16\times16$ and a temporal tubelet size of 2.
Multi-group patch masking is combined with temporal-slice masking during video latent prediction, while the target encoder is updated by exponential moving average with a momentum coefficient of 0.99925. Pretraining uses AdamW with a starting learning rate of $5.25\times10^{-4}$ and a global batch size of 4096.
After pretraining, the EMA target encoder is retained as the visual representation encoder for representation probing and downstream end-to-end planning.

The action-free pretraining data are constructed from the OpenDV-YouTube subset introduced by \citep{yang2024generalized}. Each raw video is segmented using a 4-second sliding window with a temporal stride of 0.5 seconds, resulting in a 3.5-second overlap between consecutive clips. Each clip is subsequently sampled at 2\,Hz to obtain eight input frames. For data-scaling experiments, we construct four pretraining subsets containing 0.5M, 1M, 3M, and 6M video clips. The 6M-clip subset is derived from approximately 1157 hours of raw driving videos, while the 0.5M-, 1M-, and 3M-clip subsets span 72, 131, and 587 raw driving videos, respectively. The model architecture, optimization configuration, and training budget are kept
identical across all scales, providing a controlled setting where the amount of pretraining data is the only experimental variable. Table~\ref{tab:scaledrive_video_latent_training} summarizes the detailed pretraining configuration.

\subsection{Downstream planning on NAVSIM.}
\label{sec:planning}
We keep the pretrained video encoder frozen during downstream planning training. The visual input consists of four history frames from the front camera and the current frame from each surround camera, with all images resized to $256 \times 256$ pixels. We evaluate two planner variants. The single-modal variant uses a single learnable trajectory query to predict one trajectory. The multimodal variant uses 64 learnable proposal queries to predict 64 candidate trajectories. Each predicted trajectory comprises eight poses over a 4-second planning horizon. The single-modal and multimodal variants achieve PDMS scores of 89.0 and 93.8 on NAVSIM, respectively. The single-modal planner is trained solely with an L1 trajectory loss:
\begin{equation}
    \mathcal{L}_{\mathrm{single}} = \mathcal{L}_{\mathrm{traj}}.
\end{equation}
The multimodal planner additionally uses a score loss to supervise candidate scoring:
\begin{equation}
    \mathcal{L}_{\mathrm{multi}}
    = \mathcal{L}_{\mathrm{traj}}
    + \mathcal{L}_{\mathrm{score}}
\end{equation}
Both variants are trained on 85K NAVSIM navtrain trainset using AdamW with a base learning rate of $2.0 \times 10^{-4}$. The single-modal model is trained with 100 epochs, while the multimodal model is trained with 20 epochs. Training is performed on eight NVIDIA A800 GPUs with a global batch size of 64. Table~\ref{tab:scaledrive_planner_training} summarizes the training configurations.

\begin{table}[t]
\centering
\caption{
\textbf{ScaleDrive planning training configuration on NAVSIM.}
}
\label{tab:scaledrive_planner_training}
\setlength{\tabcolsep}{6pt}
\begin{tabular}{lll}
\toprule
\textbf{Category} & \textbf{Hyperparameter} & \textbf{Value} \\
\midrule

\multirow{6}{*}{Model}
& Backbone & ViT-Large with 3D RoPE \\
& Resolution & $256 \times 256$ \\
& Front-view frames & 4 \\
& Surround-view frames & 1 \\
& Patch size & $16 \times 16$ \\
& Tubelet size & 2 \\

\midrule

\multirow{2}{*}{Planning}
& Predicted poses & 8 \\
& Planning horizon & 4\,s \\

\midrule

\multirow{5}{*}{Training}
& Trainable parameters & 17.575M \\
& Optimizer & AdamW \\
& Learning rate & $2.0 \times 10^{-4}$ \\
& Number of GPUs & 8 $\times$ NVIDIA A800 \\
& Global batch size & 64 \\

\midrule

\multirow{2}{*}{Data}
& Dataset & NAVSIM navtrain \\
& Dataset size & 85K \\

\bottomrule
\end{tabular}
\end{table}

\subsection{Cross-Planner Configurations}
\label{app:cross_planner_configs}

We evaluate video representation transfer with four downstream planners: LTF~\cite{chitta2022transfuser}, DiffusionDrive~\cite{liao2025diffusiondrive}, DrivoR~\cite{kirby2026driving}, and ChainFlow~\cite{wang2026chainflow}. All configurations use multi-view images. For each planner architecture, we compare its original baseline with V-JEPA2 and ScaleDrive variants.
Table~\ref{tab:cross_planner_configs} summarizes the native
architecture and the shared video representation integration strategy. For each planner, we retain the training losses, optimizer, learning-rate schedule, and training data used in its original paper. These settings are kept identical across the original baseline, V-JEPA2, and ScaleDrive variants within each planner.

\begin{table}[t]
\centering
\small
\setlength{\tabcolsep}{5pt}
\caption{\textbf{Cross-planner configurations.}~V-JEPA2 and ScaleDrive use the same integration strategy
within each planner.}
\label{tab:cross_planner_configs}
\begin{tabular}{lccc}
\toprule
\textbf{Planner} & \textbf{Backbone}
& \textbf{Decoder} & \textbf{Video representation integration} \\
\midrule
LTF & TransFuser & Regression
& Attention keys and values \\
DiffusionDrive & TransFuser & Diffusion
& Attention keys and values \\
DrivoR & DINOv2 & Regression
& Front-view feature replacement \\
ChainFlow & DINOv2 & Autoregressive
& Front-view feature replacement \\
\bottomrule
\end{tabular}
\end{table}

\paragraph{Additional feature fusion: LTF and DiffusionDrive.}
LTF and DiffusionDrive use a Latent TransFuser backbone to encode
multi-view images into a BEV representation, followed by regression and diffusion trajectory decoders. For both planners, the video-enhanced variants retain the native backbone features and introduce video representations as additional attention keys and values in the planner. This allows the planner to attend to the video representation alongside its original visual features.

\paragraph{Front-view feature replacement: DrivoR and ChainFlow.}
DrivoR and ChainFlow use a fine-tuned DINOv2 backbone to encode multi-view
images, with regression and autoregressive trajectory decoders,
respectively. For these planners, we replace only the front-view DINOv2 features with corresponding video representations. For these two planners, video features replace the native front-view representation rather than serving as an additional feature branch.

\subsection{Cross-Benchmark Configurations}
\label{app:cross_benchmark_configs}

\paragraph{HUGSIM.}~We directly evaluate the LTF and ChainFlow planners trained with the NAVSIM training set, including their original, V-JEPA2, and ScaleDrive variants, on HUGSIM without additional training or fine-tuning. All variants are evaluated under the same HUGSIM protocol. This setting assesses the cross-benchmark generalization of the trained driving policies without adaptation to the target environment.

\paragraph{Bench2Drive.}~We evaluate ScaleDrive on Bench2Drive using TFv6~\cite{nguyen2026lead} as the downstream planner. The pretrained ScaleDrive encoder remains frozen and processes four-frame clips at a resolution of $256 \times 256$. TFv6 uses RegNetY-032 backbones for both image and LiDAR inputs, with camera images resized to $384 \times 384$. We train the planner on the LEAD dataset for 30 epochs using AdamW with a learning rate of $3.0 \times 10^{-4}$, a weight decay of 0.01, and a global batch size of 48. Table~\ref{tab:bench2drive_training} summarizes the detailed training configuration.

\begin{table}[t]
\centering
\caption{
\textbf{Training configuration on Bench2Drive.}
}
\label{tab:bench2drive_training}
\setlength{\tabcolsep}{6pt}
\begin{tabular}{lll}
\toprule
\textbf{Category} & \textbf{Hyperparameter} & \textbf{Value} \\
\midrule

\multirow{5}{*}{Model}
& Backbone & ViT-Large with 3D RoPE \\
& Resolution & $256 \times 256$ \\
& Frames per clip & 4 \\
& Patch size & $16 \times 16$ \\
& Tubelet size & 2 \\

\midrule

\multirow{4}{*}{Planning}
& Planner architecture & TFv6 \\
& Image backbone & RegNetY-032 \\
& LiDAR backbone & RegNetY-032 \\
& Input resolution per camera & $384 \times 384$ \\

\midrule

\multirow{5}{*}{Training}
& Optimizer & AdamW \\
& Learning rate & $3.0 \times 10^{-4}$ \\
& Weight decay & 0.01 \\
& Global batch size & 48 \\
& Training epochs & 30 \\

\midrule

Data
& Dataset & LEAD \\
& Dataset size & 960K \\

\bottomrule
\end{tabular}
\end{table}

\begin{table}[t]
\centering
\caption{
\textbf{Training configuration for ScaleDrive representation probing.}
}
\label{tab:probe-training}
\small
\setlength{\tabcolsep}{4pt}
\renewcommand{\arraystretch}{1.10}
\begin{tabular}{llll}
\toprule
\textbf{Category} &
\textbf{Hyperparameter} &
\textbf{Depth Estimation} &
\textbf{Ego-Motion Estimation} \\
\midrule

\multirow{4}{*}{Model}
& Backbone
& ViT-Large with 3D RoPE
& ViT-Large with 3D RoPE \\

& Resolution
& $256\times256$
& $256\times256$ \\

& Tubelet size
& 2
& 2 \\

& Encoder token grid
& $2\times16\times16$
& $2\times16\times16$ \\

\midrule

\multirow{4}{*}{Probe}
& Prediction target
& Depth map
& Waypoints \\

& Feature aggregation
& Temporal mean pooling
& 3 learnable queries \\

& Probe architecture
& BatchNorm + Linear $1024\rightarrow1$
& Cross-attention block \\

& Output head
& Bilinear upsampling + Softplus
& LayerNorm + MLP $1024\rightarrow1024\rightarrow2$ \\

\midrule

\multirow{5}{*}{Training}
& Optimizer
& AdamW
& AdamW \\

& Learning rate
& $3.0\times10^{-4}$
& $3.0\times10^{-4}$ \\
& Weight decay
& 0.05
& 0.05 \\
& Global batch size
& 16
& 16 \\
& Training epochs
& 10
& 10 \\
\midrule
Data
& Dataset
& KITTI
& NAVSIM \\
\bottomrule
\end{tabular}
\end{table}

\subsection{ScaleDrive representation probing}
\label{app:frozen-probes}

The pretrained visual representation is evaluated through two complementary probing tasks: monocular depth estimation for spatial geometry and inverse-dynamics ego-motion estimation for temporal dynamics. Both tasks use the same pretrained ViT-Large video encoder. The encoder remains frozen throughout probe training, while only the task-specific probe is optimized.

\textbf{Depth Estimation.}
We evaluate spatial representation quality through monocular depth estimation on KITTI. The frozen spatiotemporal features are first aggregated along the temporal dimension by mean pooling. The resulting spatial features are passed through a shallow depth head consisting of BatchNorm and a linear projection, followed by bilinear upsampling and a Softplus activation to produce dense depth predictions at the input resolution. Table~\ref{tab:probe-training} summarizes the probe architectures and optimization settings.

\textbf{Ego-Motion Estimation.}
We evaluate temporal dynamics through inverse-dynamics ego-motion estimation on NAVSIM. Learnable queries aggregate the frozen spatiotemporal tokens through a cross-attention block. The resulting query features are normalized and decoded by an MLP to predict
the ego trajectory associated with the observed video sequence. This probe assesses how readily motion information can be extracted from the frozen video representation.

\begin{table}[t]
\centering
\caption{
\textbf{Comparison with NAVSIM navtrain dataset 
pretraining.}
}
\label{tab:navsim_pretrain_comparison}
\setlength{\tabcolsep}{7pt}
\begin{tabular}{l|ccccc}
\toprule
Pretraining Data & NC $\uparrow$ & DAC $\uparrow$ & EP $\uparrow$ & TTC $\uparrow$ & PDMS $\uparrow$ \\
\midrule
w/o. Pretrain & 99.0 & 98.0 & 88.4 & 96.5 & 92.4 \\
\midrule
NAVSIM
& 98.6 & 98.1 & 90.4 & 95.2 & 92.7 \\
OpenDV
& \textbf{99.3} & \textbf{98.4} & \textbf{90.6} & \textbf{97.0} & \textbf{93.8} \\
\bottomrule
\end{tabular}
\end{table}

\begin{table}[!t]
\centering
\caption{
\textbf{Detailed data-scaling results on NAVSIM.}
}
\label{tab:detailed_scaling}
\setlength{\tabcolsep}{8pt}
\begin{tabular}{l|c|ccccc}
\toprule
Method & Pretraining Data & NC $\uparrow$ & DAC $\uparrow$ & EP $\uparrow$ & TTC $\uparrow$ & PDMS $\uparrow$ \\
\midrule
V-JEPA2 & -- & 99.03 & 98.18 & 88.62 & 96.10 & 92.40 \\
\midrule
\multirow{4}{*}{ScaleDrive} & 0.5M & 98.99 & 97.92 & 89.65 & 96.12 & 92.77  \\
 & 1M & 99.10 & 97.99 & 89.78 & 96.57 & 93.05 \\
 & 3M & 99.14 & 98.10 & 90.32 & 96.51 & 93.30 \\
 & 6M & \textbf{99.16} & \textbf{98.28} & \textbf{90.67} & \textbf{96.58} & \textbf{93.61} \\
\bottomrule
\end{tabular}
\end{table}

\begin{table}[!h]
\centering

\caption{
\textbf{Ablation on temporal-slice masking.}~All configurations share the same pretraining data and training settings. Ego ADE/FDE are measured with the frozen-encoder ego-motion probe.
}
\label{tab:temporal_slice_ablation}
\setlength{\tabcolsep}{10pt}
\begin{tabular}{l|cccc}
\toprule
Masking & PDMS$\uparrow$ & EPDMS$\uparrow$  & Ego ADE$\downarrow$ & Ego FDE$\downarrow$ \\
\midrule
Patch only & 93.61 & 89.96 & 0.193 & 0.284 \\
Patch + Temporal-slice & \textbf{93.79} & \textbf{90.48} & \textbf{0.172} & \textbf{0.246} \\
\bottomrule
\end{tabular}
\end{table}

\section{More Experiments}

\paragraph{Comparison with NAVSIM pretraining.}

To further disentangle the effect of large-scale action-free pretraining, ScaleDrive is additionally pretrained on NAVSIM driving videos under the same model architecture, optimization configuration, and downstream planning protocol. This comparison directly contrasts broad open-world driving-video pretraining with benchmark-specific driving data. As shown in Table~\ref{tab:navsim_pretrain_comparison}, pretraining on NAVSIM achieves a PDMS of 92.7, whereas scaling pretraining to 6M OpenDV clips improves overall planning performance. These results show that the benefit of ScaleDrive does not depend on pretraining with downstream benchmark data, but instead increases with large-scale action-free driving-video pretraining.

\paragraph{Detailed scaling results}~Table~\ref{tab:detailed_scaling} provides the complete metric breakdown across different pretraining data scales. As the amount of pretraining data
increases from 0.5M to 6M clips, PDMS improves monotonically from 92.77 to 93.61. The improvement is accompanied by a consistent increase in EP and generally stronger TTC, while NC and DAC remain at a similarly high level. These detailed results further show that the planning benefit from scaling action-free video pretraining is reflected across multiple components of the closed-loop evaluation rather than a single aggregate metric.

\paragraph{Ablations on temporal-slice masking.}
Table~\ref{tab:temporal_slice_ablation} compares patch-only masking with the full
patch- and temporal-slice masking under identical pretraining data, backbone,
training steps, and optimization configuration, evaluated with the same
downstream planner. Beyond closed-loop planning (PDMS), we report the
frozen-encoder ego-motion probe, since Proposition~\ref{prop:temporal_recovery}
concerns temporal state inference rather than planning performance alone.

\begin{figure}[!h]
\centering
\includegraphics[width=0.97\linewidth]{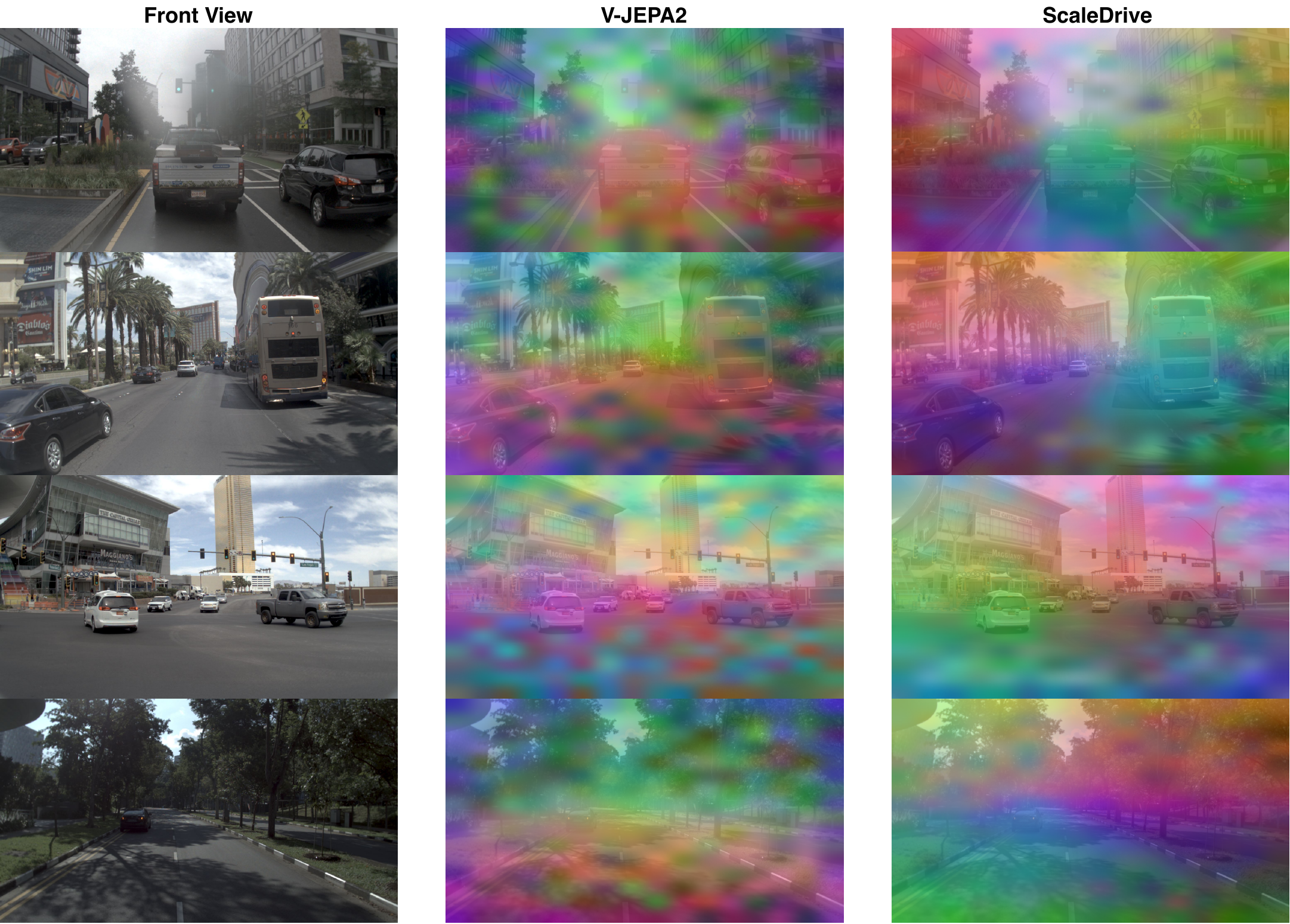}
\vspace{-2mm}
\caption{
\textbf{Visual representation comparison.}
}
\label{fig:supp3}
\vspace{-4mm}
\end{figure}

\begin{figure}[t]
\centering
\includegraphics[width=\linewidth]{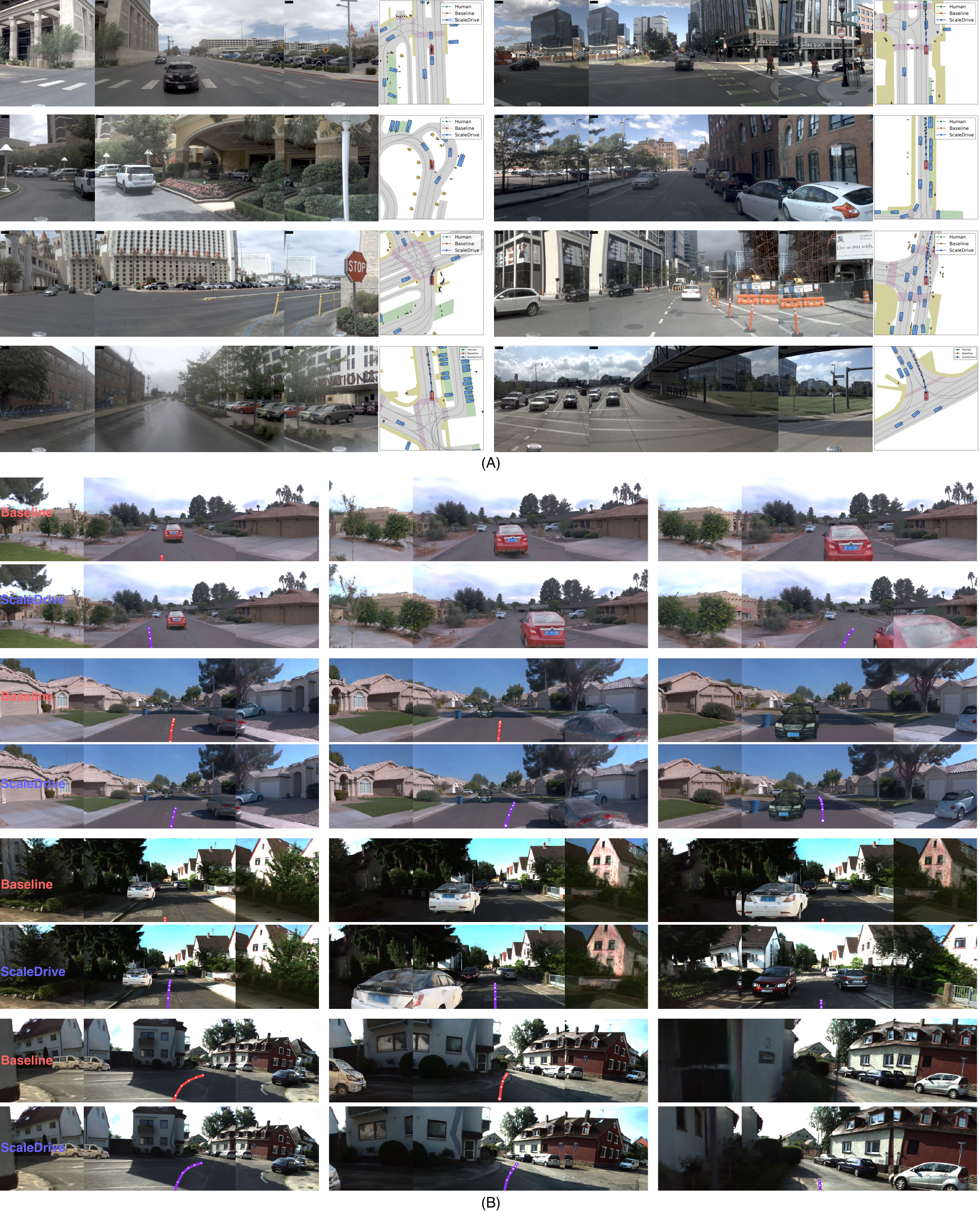}
\vspace{-7mm}
\caption{
\textbf{{ScaleDrive planning results.}(A)~NAVSIM and (B)~HUGSIM.}
}
\label{fig:supp1}
\vspace{-3mm}
\end{figure}

\paragraph{Additional qualitative visualizations}~Additional qualitative results are provided to complement the visual analyses in
the main paper. Fig.~\ref{fig:supp3} presents more PCA visualizations of the learned visual representations across diverse driving scenes, providing a broader view of the feature patterns produced by ScaleDrive. Fig.~\ref{fig:supp1} further presents additional planning examples across different driving environments in NAVSIM and HUGSIM.

\section{Limitations}
ScaleDrive focuses on learning transferable representations from monocular driving videos and does not explicitly model cross-view correspondences during pretraining. Although the frozen representations can be integrated into multi-camera planners, extending latent prediction to synchronized views may further improve spatial consistency across cameras. In addition, our experiments use a fixed temporal sampling rate and clip length, leaving adaptive temporal sampling and longer-context pretraining unexplored.

\end{document}